\documentclass[10pt,twocolumn,letterpaper]{article}

\usepackage[pagenumbers]{cvpr} 

\usepackage{graphicx}
\usepackage{amsmath}
\usepackage{amssymb}
\usepackage{booktabs}

\usepackage{xcolor}
\usepackage{amsmath}
\usepackage{amssymb}
\usepackage{booktabs}
\usepackage{graphicx}
\usepackage{bbold}
\usepackage{bbding}
\usepackage{multirow}
\usepackage{pifont}
\usepackage{subcaption}
\usepackage{blindtext}
\usepackage{array}
\usepackage{wrapfig}
\newcommand{\PreserveBackslash}[1]{\let\temp=\\#1\let\\=\temp}
\newcolumntype{C}[1]{>{\PreserveBackslash\centering}p{#1}}
\newcolumntype{R}[1]{>{\PreserveBackslash\raggedleft}p{#1}}
\newcolumntype{L}[1]{>{\PreserveBackslash\raggedright}p{#1}}
\allowdisplaybreaks
\newcommand{\CLS}{\textbf{\texttt{CLS}}\@\xspace}
\newcommand{\ua}{$\uparrow$}
\newcommand{\da}{$\downarrow$}

\usepackage{dblfloatfix}

\definecolor{cvprblue}{rgb}{0.21,0.49,0.74}
\usepackage[pagebackref,breaklinks,colorlinks,allcolors=cvprblue]{hyperref}

\def\paperID{11232} 
\def\confName{CVPR}
\def\confYear{2025}

\title{Disentangling Visual Transformers: Patch-level Interpretability for Image Classification}

\author{Guillaume Jeanneret$^{*,\ddag}$, Lo\"ic Simon$^\dagger$, Fr\'ed\'eric Jurie$^\dagger$ \\
$^*$ISIR - Sorbonne University \\
$^\dagger$Normandy University, ENSICAEN, UNICAEN, CNRS, GREYC}

\begin{document}
\maketitle

\begin{abstract}
Visual transformers have achieved remarkable performance in image classification tasks, but this performance gain has come at the cost of interpretability. One of the main obstacles to the interpretation of transformers is the self-attention mechanism, which mixes visual information across the whole image in a complex way. In this paper, we propose Hindered Transformer (HiT), a novel interpretable by design architecture inspired by visual transformers. Our proposed architecture rethinks the design of transformers to better disentangle patch influences at the classification stage. Ultimately, HiT can be interpreted as a linear combination of patch-level information. We show that the advantages of our approach in terms of explicability come with a reasonable trade-off in performance, making it an attractive alternative for applications where interpretability is paramount.
\end{abstract}
\let\thefootnote\relax\footnotetext{$^\ddag$ Work done at GREYC Laboratory, now at ISIR Laboratory}

\section{Introduction}
Deep learning architectures have achieved remarkable breakthroughs in domains such text~\cite{Dubey2024TheL3}, vision~\cite{oquab2024dinov,esser2024scaling}, or multimodal tasks~\cite{NEURIPS2023_6dcf277e}, prompting widespread interest in their application to real-world problems~\cite{li2023llavamed}. However, as these models are increasingly used in high-stakes scenarios, understanding their decision-making process becomes crucial to ensure their decisions are grounded in meaningful variables rather than spurious correlations~\cite{Jeanneret_2022_ACCV,augustin2022diffusion}. This necessity has driven the development of trustworthy and interpretable architectures.

Interpretable-by-Design (ID) architectures~\cite{NEURIPS2019_adf7ee2d} aim to inherently explain their decision-making process, eliminating the need for external interpretability tools. Ideally, these architectures maintain comparable performance to traditional black-box methods while providing insights into their internal workings. Despite their promise, existing ID architectures largely rely on convolutional neural networks (CNNs) as feature extractors~\cite{wang2023learning,Boehle2022CVPR,pmlr-v119-koh20a,oikarinen2023labelfree}. Given the recent success of transformer-based models~\cite{dosovitskiy2021an}, transitioning to interpretable transformers is a logical next step.

Although Vision Transformers (ViTs)~\cite{dosovitskiy2021an} have demonstrated superior performance in computer vision tasks, the literature on their explainability and interpretability remains sparse. While attention maps are sometimes considered interpretable, many studies~\cite{serrano-smith-2019-attention,Jain2019AttentionIN} argue that they provide little to no insight into the actual decision-making process. 
We concur that attention maps offer only partial and insufficient cues, and instead route towards ID architectures.

In this paper, we address this gap by introducing a novel interpretable transformer-like architecture. Our proposed model, the Hindered Transformer (HiT), advances the understanding of ViTs by analyzing the flow of individual image patches and decomposing the classification token (\CLS) into contributions from each individual token. By constructing predictions as the sum of these token contributions, HiT achieves interpretability without relying on external methods~\cite{alain2016understanding} or gradient-based approaches~\cite{simonyan2013deep}, classifying it as an ID method.

We summarize our contributions as follows: i) We propose the Hindered Transformer (HiT) backbone, a variant of vision transformers that is inherently interpretable. ii) We empirically validate HiT on six datasets -- ImageNet~\cite{deng2009imagenet}, CUB 2011~\cite{WahCUB_200_2011}, Stanford Dogs~\cite{KhoslaYaoJayadevaprakashFeiFei_FGVC2011}, Stanford Cars~\cite{krause20133d}, FGVC-Aircraft~\cite{maji13fine-grained}, and Oxford-IIIT Pets~\cite{parkhi12a} -- demonstrating its interpretability both quantitatively and qualitatively, outperforming recent ID transformer-based models and post-hoc methods.


Our code and pretrained models are found on \href{https://github.com/guillaumejs2403/HiT}{GitHub}

\section{Related Work}

The performance of neural networks on computer vision tasks is well-established, but the complexity of these models can make them difficult to understand. This lack of transparency is problematic and has motivated the scientific community to develop methods for making neural networks more interpretable. There are two main approaches to this problem: post-hoc methods, which seek to analyze an already-trained model, and interpretable by design architectures, which aim to create models whose decision-making processes are inherently transparent.

Many post-hoc methods have been proposed in the literature, including counterfactual explanations~\cite{augustin2022diffusion,Jeanneret_2023_CVPR,zemni2023octet}, saliency maps~\cite{Wang_2023_ICCV,Jalwana_2021_CVPR,8237336,Petsiuk2018rise}, and model distillation~\cite{Ge_2021_CVPR,tan2018learning}. 
Closer to our work, a few attempts have been made to explain a ViT architecture via post-hoc algorithms. 
For instance, Abnar and Zuidema~\cite{abnar2020quantifying} proposed to compute a score per token by recursively propagating the attention maps in a top-bottom approach. 
In addition, some methods extended the LRP~\cite{bach2015pixel} paradigm to include the attention heads~\cite{chefer2021generic,Chefer_2021_CVPR}. 
The previous methods leverage the mechanisms of transformers to estimate the individual contribution of each token. 
While these approaches provide insights into trained models, they often rely on external tools or post-processing, which can limit their generalizability and integration into the model itself.

 
The field of interpretable by design architectures is diverse, as there is no single approach to explaining the complex behaviors neural networks. While many methods have been proposed, there has been a recent focus on prototypical part networks, such as ProtoPNets introduced by Chen~\etal~\cite{NEURIPS2019_adf7ee2d}. ProtoPNets computes class predictions based on the distances between patches of the final feature map and some prototypes, which can be visualized. 
Additionally, there have been many variations of ProtoPNets proposed in the literature~\cite{wang2023learning,Nauta_2023_CVPR,nauta2021looks,carmichael2024pixel,bontempelli2023conceptlevel,ukai2023this, rymarczyk2021interpretable,Wang_2021_ICCV,nauta2021neural,donnelly2022deformable,Dai_2017_ICCV,hase2019interpretable}. These variations aim to refine ProtoPNets to enhance performance and interpretability but primarily focus on convolutional neural networks (CNNs)~\cite{7780459,Simonyan15}, limiting their application to modern transformer architectures.

In addition to ProtoPNets, there are other methods that use alternative forms of prototypes. 
For example, PDiscoNet~\cite{van2023pdisconet} automatically detects parts of objects and uses them for the final classification. Similarly, BagNet~\cite{brendel2018approximating} mimics the Bag-of-Features approach to understand the decision-making process. 
Concept Bottleneck Models~\cite{pmlr-v119-koh20a,oikarinen2023labelfree} use concepts to explain their decisions.
Finally, a family of networks propose interpretable layers, such as B-cos networks~\cite{Boehle2022CVPR}, that learns an easily interpretable input-dependent linear transformation. 
The work of Zhang~\etal~\cite{zhang2018interpretable} proposed convolutional interpretable layers. 
Finally, some works use some variation of decoupled networks~\cite{liang2020training,Li2019InterpretableNN,shen2021interpretable} to highlight what a filter is looking at.


Concerning visual transformers, several recent works have attempted to make transformers more interpretable by modifying their architecture. Some papers try to push the boundaries to include transformer-based heads~\cite{hong2024concept,Paul2023ASI,rigotti2022attentionbased,pmlr-v162-kim22g} for prototype interpretability. However, they only include a single self-attention mechanism on top of a CNN backbone.  ProtoPFormer~\cite{Xue2022ProtoPFormerCO} suggests including a ViT backbone in the prototype setup. To do this, ProtoPFormer includes a prototype layer on top of both the classification and image tokens. 
While innovative, this architecture does not ensure that the tokens contain purely local information due to the increased receptive fields of tokens when combining classification and image tokens~\cite{raghu2021do}. B-cos networks V2~\cite{Boehle2024TPAMI} use the same rational as the original B-cos~\cite{Boehle2022CVPR} approach, but they extend it to transformers. In a few words, B-cos networks summarize their inner workings as a single linear function, creating the attribution map by simply multiplying the input and the summarized network. However, this extension still relies on convolutional layers as the initial feature extraction stage, limiting its interpretability scope. 
Finally, A-ViT~\cite{yin2022avit} was originally designed for faster inference by removing tokens at certain layers. Despite this, this mechanism serves as an interpretable system, as the most important tokens are retained until the last layer.  This approach, however, focuses on efficiency rather than interpretability as its primary goal. 


In contrast with this literature, we propose an architecture interpretable by design, by adapting the main building blocks of vision transformers to disentangle the contributions of each image patch. This enables us to compute the salient regions of the image without the need for any non-traditional training or invasive methods. 
Unlike other approaches, we avoid relying on the attention mechanism of the multi-head attention (MHA) block to produce saliency maps. Instead, our network inherently generates these maps as part of its decision-making process.

\section{Methodology}


In this section, we present our proposed approach and the rationale behind it. First, in \S\ref{sec:prelim-hit}, we present the preliminaries for the multi-head attention mechanism and ViTs. Next, in \S\ref{sec:mha} we will show that attention and multi-head attention output can be decomposed into the individual contributions of the inputs. Finally, in \S\ref{sec:presenting-hit} we present our novel architecture, the Hindered Transformer (HiT). The core of our method is to minimise the mixing of patch-level information, which allows us to express the classification token (\CLS) in ViTs as the sum of individual tokens, a direct result of \S\ref{sec:mha}. In other words, this simplification allows us to check the contribution of each token.

\subsection{Preliminaries: Transformers, ViTs and Notations}\label{sec:prelim-hit}

The transformer architecture is built upon the Scaled Dot-Product Attention operation~\cite{vaswani2017attention}, commonly referred to as \emph{attention}. Given a query token sequence ${x}^q \in \mathbb{R}^{L_q \times d_{model}}$ and a target sequence (or key-value sequence) ${x}^t \in \mathbb{R}^{L_t \times d_{model}}$, where $L_q$ and $L_t$ are their respective sequence lengths and $d_{model}$ is the token dimension, the attention mechanism is computed as follows:
\begin{equation}\label{eq:attn}
\begin{split}
    Q &= x^q\,W_Q + b_Q \\
    K &= x^t\,W_K + b_K \\
    V &= x^t\,W_V + b_V \\
    A(x^q, x^t) &= softmax\left(\frac{QK^T}{\sqrt{d_k}}\right)V
\end{split}
\end{equation}
where the output is a sequence of the same length as $x^q$, $d_k$ is the dimension of the linear transformations, and $W_i\in\mathbb{R}^{d_{model}\times d_k}$ and $b_i\in\mathbb{R}^{d_k}$ are the weights of the linear projection $i\in\{Q, K, V\}$. 
In addition, Vaswani \textit{et al.}~\cite{vaswani2017attention} proposed to compute the attention mechanism $h$ times in parallel, setting $d_k = d_{model} / h$ for each individual attention operation. The resulting vectors of each individual attention, formally called heads, are concatenated and linearly post-processed to obtain the final result. This operation is called multi-head attention, and it is described as follows:
\begin{equation}
    \small MHA(x^q, x^t) = \underbrace{[A^1(x^q, x^t); ...; A^h(x^q, x^t)]}_{\texttt{Concatenate $h$ times}}W_o + b_o,
\end{equation}
with $A^i$ being the $i^{th}$ attention mechanism in the MHA, and $W_o\in\mathbb{R}^{d_{model}\times d_{model}}$ and $b_o\in\mathbb{R}^{d_{model}}$ the linear transformation parameters. 

In computer vision, to incorporate image data into this sequence-based formulation, the ViT first partitions the input image into $N^2$ equal-sized patches and linearly projects them to create the patch token sequence\footnote{For the rest of the paper, we will use the terms token and patch interchangeably, referring to the image patch tokens.}.
Additionally, following standard practice, a learnable classification token \CLS is prepended to the patch sequence. Finally, each patch token is summed with a positional embedding to encode its spatial location within the image. For the remainder of the paper, the sequence $x \in \mathbb{R}^{(N^2 + 1) \times d_{model}}$ denotes the concatenation of the patch tokens and the \CLS token, where $x[0]$ corresponds to the \CLS token.

The main ViT block builds on the MHA operation, followed by a token-wise MLP block, as in text-based transformers. 
Formally, given a set of patches $x_l$ at layer $l$, the ViT block first computes a globalized set of tokens using the MHA block. 
The resulting output is summed with a skip connection. 
Then, the output is fed into a token-wise MLP to post-process each token, followed, again, by a skip connection.
This block is summarized as follows
\begin{equation}\label{eq:vit-block}
\begin{split}
    x_{l}' &= x_l + MHA(x_l, x_l) \\
    x_{l+1} &= x_l' + MLP(x_l')
\end{split}
\end{equation}
Note that before the MHA and MLP blocks, a LayerNorm~\cite{ba2016layer} operation is applied to the data sequence, but for simplicity, we omit this operation.
Finally, the \CLS token is fed into a LayerNorm followed by a linear classifier to produce the logits of the classification task.

\subsection{Multi-Head Attention and Patch Mixing in Transformers}\label{sec:mha}

In this section, we aim to decompose the MHA operation to demonstrate that it is possible to retrieve the individual contributions of each token. In this way, we aim to lay the foundation for our architecture, which is described in the next section. 

Let's start by focusing on the attention operation (Eq.~\ref{eq:attn}). Since we will focus on the \CLS token later, and to simplify the analysis, let's assume that the query sequence has length $L_q=1$. Consequently, the attention mechanism can be rewritten as
\begin{equation}\label{eq:attn-decomp}
    A(x^q, x^t) = \sum_{v\in x^t} s(v, x^q, x^t)(v\,W_V + b_v),
\end{equation}
where $s(v, x^q, x^t) \in \mathbb{R}$ is the contribution factor of $v \in x^t$, computed via the softmax operation in Eq.~\ref{eq:attn}. Here, Eq.~\ref{eq:attn-decomp} shows that we can decompose the attention mechanism into separately processed patches - each patch $v$ in $x^t$ adds $s(v, x^q, x^t)(v\,W_V + b_v)$. Accordingly, if $x^t$ contains purely local information, the output of the attention is \emph{a weighted sum of local data}. Here, we define whether a token contains local information as the product of processing a single patch without the influence of other datapoints.  

To continue, we incorporate the previous observation into multi-head attention and verify that we can still unroll this operation into a \emph{sum of separate vectors}. One might be concerned that the concatenation-linear operation will mix each token. However, we argue that the result is still valid, since concatenating and linearly transforming the resulting vector is equivalent to linearly transforming each head and adding them together. 
Formally, by denoting $W_v^i$ and $b_v^i$ as the weights of the linear transformation generating the value sequence of $i^{th}$ head, and breaking apart $W_o$ into $h$ separate matrices, $W_o = [W_o^1; W_o^2; ...; W_o^h]$, with $W_o^i\in\mathbb{R}^{d_k\times d_{model}}$, then, the MHA becomes
\begin{equation}
\begin{split}
    MHA(x^q, x^t) &= b_o + \sum_{v\in x^t} v'(v) \\
    \text{where} \quad v'(v) &= \sum_{i=1}^h s(v, x^q, x^t) (v\,W_v^i+b_v^i) W_o^i.
\end{split}
\end{equation}
The previous result implies that we can still decompose the MHA result as the sum of vector patches, regardless of the number of heads in the MHA. So the same conclusion holds as in Eq.~\ref{eq:attn-decomp}: if the content in $x^t$ is local, then we can unravel the MHA mechanisms into \emph{local contributions}. 

\subsection{Untangling Visual Transformers}\label{sec:presenting-hit}

Unlike single MHA layers, ViTs operate on global features.
To integrate local information, these architectures use two mechanisms: the MHA layers, which spread the information within tokens, and the nonlinear MLPs, which introduce complex correlations even when applied to a linear combination of local contributions. For better explainability, it would be ideal if the classifier's decision could be expressed as a combination of information from individual patches, allowing a more interpretable understanding of how local information contributes to global predictions.

\begin{figure}
    \centering
    \includegraphics[width=0.9\linewidth]{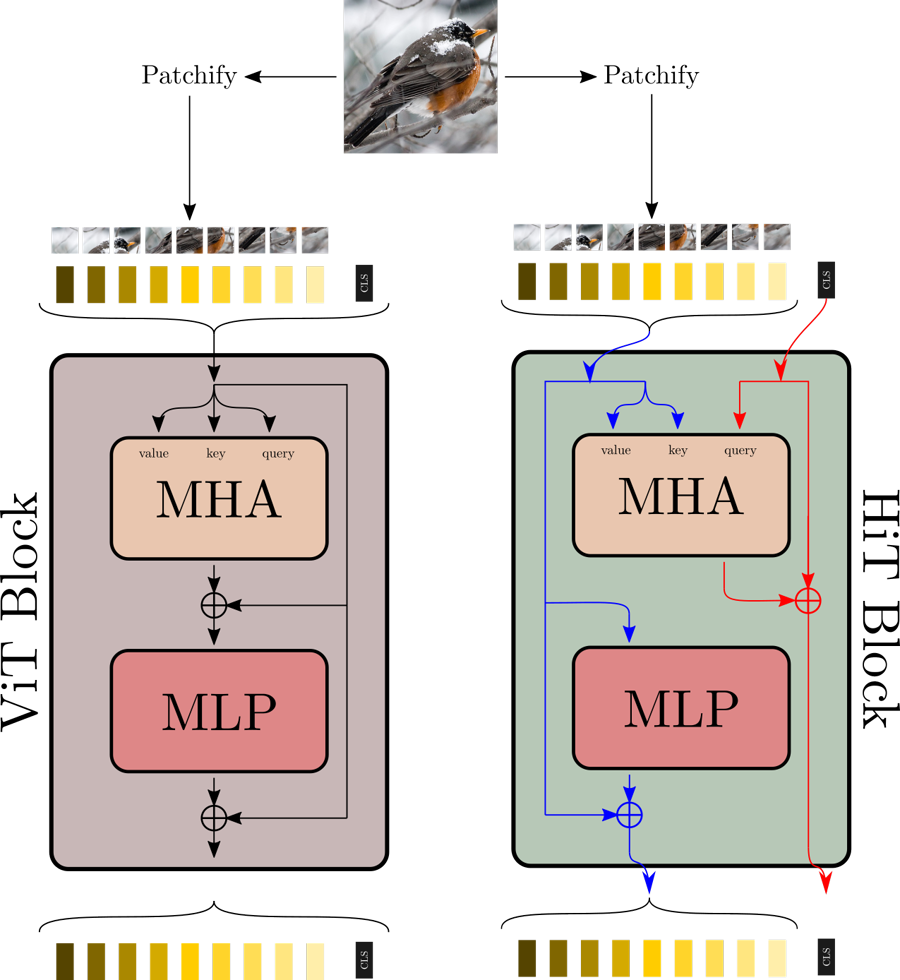}
    \caption{\textbf{ViT and HiT blocks.} While the ViT block mixes the patch data, HiT uniquely updates the \CLS via the MHA, but avoids post-processing the classification token in the MLP, allowing the \CLS to be unrolled at the last layer as individual contributions.}
    \label{fig:hit}
\end{figure}

In this section, we describe our proposed architecture: Hindered Transformer (HiT). By constraining the image tokens to contain only local information along all inference blocks, and by avoiding mixing the \CLS token, our novel method is able to partition the \CLS token into each individual patch, a direct outcome of the previous section. Fig.~\ref{fig:hit} shows the difference between the ViT block, and our block.

The first challenge is then constraining the data flow between patches.
To do so, we create an intermediate architecture that uses \CLS token $x_l[0]$ as the query in the MHA operation, and the rest of the sequence $x_l$ as the key-value input.
So, the output from the MHA is a single token that is summed to $x_l[0]$.
Then, as in ViTs, we will post-process each token in the sequence with the MLP.
Thus, the ViT update function in Eq.~\ref{eq:vit-block} is transformed to
\begin{equation}\label{eq:pat-block}
\begin{split}
    x'_l[0] &= x_l[0] + MHA(x_l[0], x_l)\\
    x'_l[1:] &= x_l[1:] \\
    x_{l+1} &= x'_l + MLP(x'_l).
\end{split}
\end{equation}

The previous model solves one problem by limiting the merging of data in local patches. 
However, processing the \CLS token through the MLP mixes the local information provided by the MHA block, as well as the value and output operations. 
Since our goal is to disentangle the data flow into individual contributions, we need to further constrain this processing.
To do this, we simply avoid updating the \CLS token through the MLP and passing it to the target sequence.
So, our block inference is
\begin{equation}\label{eq:hit-block}
\begin{split}
    x_{l+1}[0] &= x_l[0] + MHA(x_l[0], x_l[1:])\\
    x_{l+1}[1:] &= x_l[1:] + MLP(x_l[1:])
\end{split}
\end{equation}
We call the final architecture the Hindered Transformer (HiT), as we hinder the connections of the ViT. 
In a nutshell, HiT only updates the \CLS token via the MHA, while the MLP blocks update the image patches. These restrictions help to preserve purely local information in each token, while allowing the \CLS token to be unrolled. 

\begin{figure}[t]
    \centering
    \includegraphics[width=0.95\linewidth]{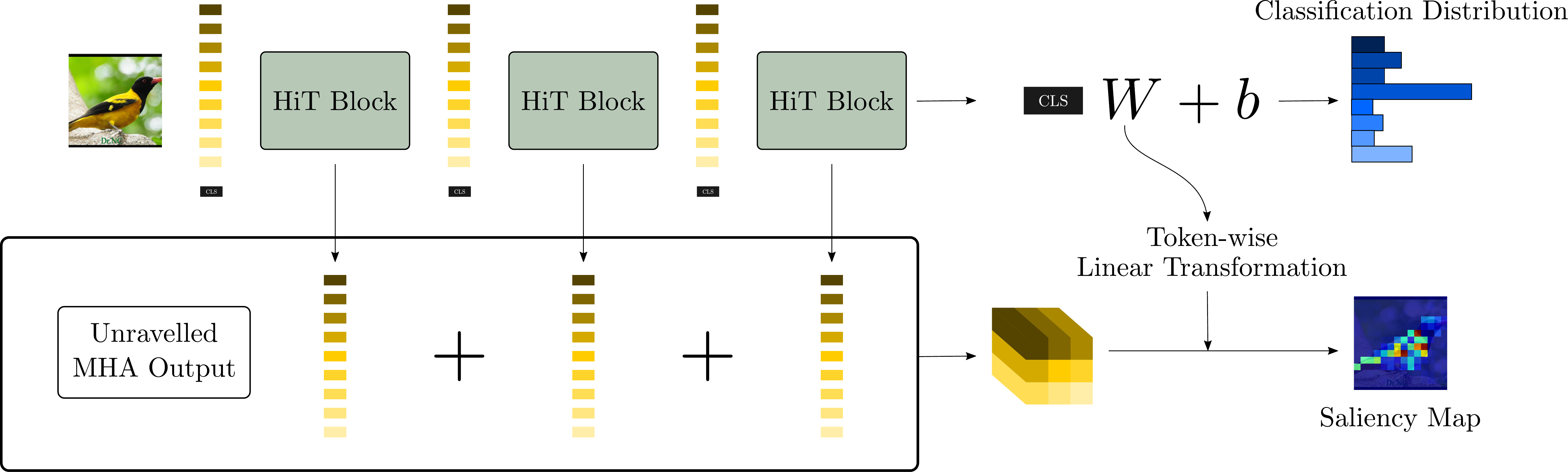}
    \caption{\textbf{Saliency Maps computation using HiT.} From the results from \S\ref{sec:mha} and the definition of our architecture, HiT enables to extract the individual contribution per token and per layer. By adding together all tokens per layer, we can rearrange the tokens in a spatial layout and use the linear layer \textit{\`a la} CAM~\cite{zhou2016learning} to extract the contribution of each token.}
    \label{fig:hit-saliency}
\end{figure}

Since the classification token is not post-processed with MLP or MHA, the final image classification is the sum of the individual tokens in all layers, as shown in \S\ref{sec:mha}. 
Therefore, the \CLS in the last layer is
\begin{equation}\label{eq:hit-cls}
\begin{split}
    x_L[0] &= x_0[0] + \sum_{l=0}^{L-1} MHA(x_l[0], x_l[1:]) \\
    &= x_0[0] + \sum_{l=0}^{L-1} \left[ b_o^l + \sum_{v\in x_l[1:]} v'_l(v) \right] \\
    &=\sum_{l=0}^{L-1} \sum_{v\in x_l[1:]} \left[v'_l(v) + \frac{b_o^l}{N^2} + \frac{x_0[0]}{LN^2} \right].
\end{split}
\end{equation}
Please note that we distribute the biases $b_o^l$ of the projection operation in the MHA head evenly to each patch $v'_l(v)$. 
In a similar fashion, we spread $x_0[0]$ into all tokens for all layers.

One advantage of this architecture is that we can easily compute saliency maps, as shown in Fig.~\ref{fig:hit-saliency}.
The double sum in Eq.~\ref{eq:hit-cls} can be decomposed as a tensor $\mathbb{R}^{L\times N^2 \times d_{model}}$, where the final image representation is the sum over the first and second dimensions, \ie the layer and token dimension, respectively.
Thus, and similarly to CAM~\cite{zhou2016learning}, to compute the regions of interest used by the model for an input image, we simply run the linear classifier on each patch to get the map.
This rationale is similar to the LRP~\cite{bach2015pixel} method in the sense that the sum of value in the saliency is equal to the output logit for that specific class.

\subsection{Token Pooling}\label{sec:pooling}


Token pooling \cite{tokenpooling}, which involves downsampling the number of tokens as one progresses through the layers, is commonly used to improve the computational efficiency of standard transformers~\cite{pan2021scalable}. This pooling technique effectively addresses the issue of representation power (as empirically demonstrated in \S~\ref{sec:perf-loss}) by expanding the receptive field of the image tokens in the deeper layers. We choose to adopt this approach due to its significant advantages.

To achieve this, we first reorganize the tokens into their spatial layout and then perform the pooling operation. However, we need to adapt the explanation generation approach to accommodate the pooled tokens. Typically, this involves using the backward operation of the pooling operator. In our case, since we rely on average pooling, which is linear, the backward operation is simply the transposed operator, which replicates each output token across the associated \(2 \times 2\) block and divides by \(4\). In other words, we distribute the importance of the pooling step equally among the contributing tokens.


\section{Experiments}



\subsection{Evaluation Protocols}

The quantitative evaluation of the saliency map quality is performed using the insertion-deletion metrics~\cite{Petsiuk2018rise}. The insertion process is performed iteratively by creating a perturbed copy of the input image by progressively adding regions from the original image. This copy is initially filled either with zeros or with a blurred version of the image. Starting with the most influential regions according to the saliency map and moving to the least influential regions, each step adds an original region in place of the perturbed one. During this process, the class probability of the perturbed image is tracked for the originally predicted class. This results in a curve for each input image, where the x-axis represents the percentage of inserted regions and the y-axis represents the probability. 
Then, the insertion metric is computed with the area under the curve (AUC) of the mean probability curve for all tested instances. 
The deletion metric works in a similar fashion, removing the original patches starting with the most influential and continuing until all patches are removed. 
If a saliency map generator accurately identifies the most relevant components influencing the model's decision, it will produce a steep increase (or decrease) in the insertion (or deletion) curve. A sharp transition in these curves confirms that the highlighted regions are indeed those utilized by the model for classification. Consequently, a reliable saliency method will achieve a higher (lower) AUC for the insertion (deletion) metric.

Insertion-deletion curves generated by different models cannot be directly compared if the models do not have the same calibration of their outputs, which is the case in these experiments. Therefore, we propose normalizing the insertion-deletion metrics by calculating the normalized AUC (nAUC). 
This normalization adjusts the mean probability curve based on the maximum and minimum values of the curve before computing the AUC.

Lastly, when assessing interpretable by-design approaches, it is necessary to evaluate the trade-off between raw performance on the task vs the gain in interpretability. Specifically, we compare the top-1 accuracy in contrast to the insertion-deletion metrics. 


As for the dataset, we evaluated HiT on six diverse image classification datasets, traditionally used for evaluating interpretability architectures: ImageNet~\cite{deng2009imagenet}, CUB-2011~\cite{WahCUB_200_2011}, Stanford Dogs~\cite{KhoslaYaoJayadevaprakashFeiFei_FGVC2011}, Stanford Cars~\cite{krause20133d}, FGVC-Aircraft~\cite{maji13fine-grained}, and Oxford-IIIT Pets~\cite{parkhi12a}. In the supplementary material, we described in-depth each dataset.

\subsection{Implementation Details}

To train HiT on ImageNet~\cite{deng2009imagenet}, we used the official DeiT3 codebase~\cite{Touvron2022DeiTIR} and followed a similar setup to their method. 
Please read the supplementary for more details on the architecture, but in short, HiT uses exactly the same depth and number of layers as ViT.
For HiT-B and HiT-S, we trained our models for 600 epochs using the AdamW optimizer~\cite{loshchilov2018decoupled} with a learning rate of $8\times10^{-4}$, a weight decay of 0.05, a batch size of 4096, 20 warm-up epochs, a cosine annealing scheduler~\cite{loshchilov2017sgdr}, and ThreeAugment~\cite{Touvron2022DeiTIR} data augmentation. 
Unlike the DeiT3 training regime, we did not use the binary cross entropy loss or any LayerDrop~\cite{Fan2020Reducing} regularization, but the traditional cross entropy with a smoothing of 0.1 and an attention dropout of 0.2.
To fine-tune HiT in the other datasets, we trained our models similarly to the ImageNet's configuration, but instead we set the batch size to 512, the number of epochs to 300 and the learning rate to $5\times10^{-5}$ for Standford Dogs and Oxford Pets, $4\times10^{-4}$ for Standford Cars and FGVC-Aircraft, and $1\times10^{-4}$ for CUB 2011.

\begin{table*}[t]
    \centering
    \scriptsize
    \begin{tabular}{C{1.7cm}|C{0.8cm}C{0.8cm}|C{0.8cm}C{0.8cm}|C{0.8cm}C{0.8cm}|C{0.8cm}C{0.8cm}|C{0.8cm}C{0.8cm}|C{0.8cm}C{0.8cm}}\toprule
        \multirow{2}{*}{Method}       & \multicolumn{2}{c|}{ImageNet} & \multicolumn{2}{c|}{CUB 2011} & \multicolumn{2}{c|}{Stanford Cars} & \multicolumn{2}{c|}{Stanford Dogs} & \multicolumn{2}{c|}{FGVC-Aircrafts} & \multicolumn{2}{c}{Oxford-IIIT Pets} \\ \cmidrule{2-13}
                      & I-Z (\ua) & D-Z (\da) & I-Z (\ua) & D-Z (\da) & I-Z (\ua) & D-Z (\da) & I-Z (\ua) & D-Z (\da) & I-Z (\ua) & D-Z (\da) & I-Z (\ua) & D-Z (\da) \\ \midrule
        HiT           & \textbf{0.57} & \textbf{0.12} & \textbf{0.36} & \textbf{0.06} & \textbf{0.66} & \textbf{0.09} & \textbf{0.64} & \textbf{0.13} & \textbf{0.60} & \textbf{0.08} & \textbf{0.67} & \textbf{0.18}\\
        HiT + Rollout & 0.40  & 0.21  & 0.47  & 0.19  & 0.49  & 0.16  & 0.50  & 0.21 & 0.52 & 0.11 & 0.58 & 0.29 \\
        HiT + GradCAM & 0.49  & 0.15  & 0.33  & 0.10  & 0.57  & 0.13  & 0.60  & 0.15 & 0.53 & 0.09 & 0.64 & 0.25 \\ \midrule
                      & I-B (\ua) & D-B (\da) & I-B (\ua) & D-B (\da) & I-B (\ua) & D-B (\da) & I-B (\ua) & D-B (\da) & I-B (\ua) & D-B (\da) & I-B (\ua) & D-B (\da) \\ \midrule
        HiT           & \textbf{0.58} & \textbf{0.23} & \textbf{0.52} & \textbf{0.14} & \textbf{0.61} & \textbf{0.20} & \textbf{0.62} & \textbf{0.27} & \textbf{0.59} & \textbf{0.17} & \textbf{0.60} & \textbf{0.26} \\
        HiT + Rollout & 0.45  & 0.31  & 0.47  & 0.19  & 0.52  & 0.28  & 0.53  & 0.35 & 0.55 & 0.2 & 0.53 & 0.35 \\
        HiT + GradCAM & 0.53  & 0.28  & 0.49  & 0.21  & 0.56  & 0.26  & 0.61  & 0.31 & 0.57 & 0.19 & 0.60 & 0.32 \\ \bottomrule
    \end{tabular}
    \caption{\textbf{HiT and Explainability methods:} We quantitatively compare HiT maps and those created by GradCAM and the modified rollout matrix (mean attention) using AUC (no normalization required here). The assessment shows that HiT maps are more faithful than those generated by GradCAM or the rollout matrix. Higher insertion is better, while lower deletion is better. I and D refers to the Insertion and Deletion metrics, respectively. Z is the zero-corrupted image, while B is the blurred corruption strategy.}
    \label{tab:interpretability-assessment}
    \vspace{-3mm}
\end{table*}

\begin{figure}[t]
    \centering
    \begin{subfigure}{0.49\textwidth}
        \centering
        \includegraphics[width=0.85\textwidth]{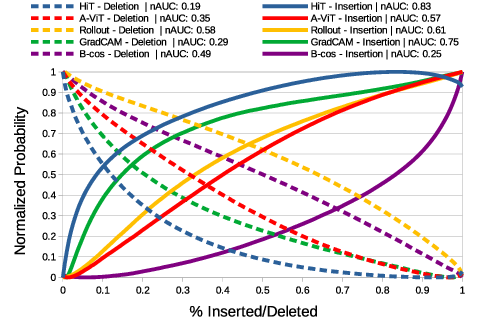}
        \caption{ImageNet}
    \end{subfigure}
    \begin{subfigure}{0.49\textwidth}
        \centering
        \includegraphics[width=.85\textwidth]{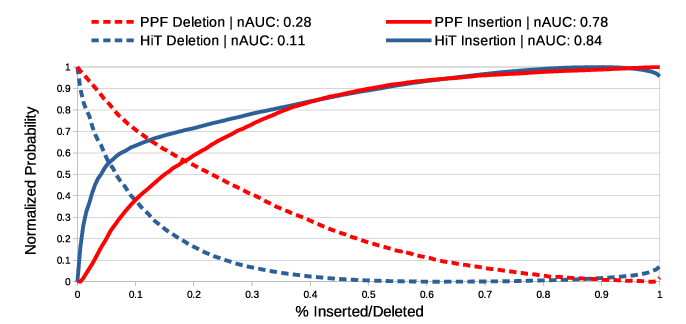}
        \caption{CUB-2011}
    \end{subfigure}
    \caption{\textbf{Interpretability comparison.} We tested whether HiT's saliency maps provide better information than ProtoPFormer's, A-ViT's maps, B-cos, the rollout attention, and GradCAM. The results indicate that our methods are indeed more interpretable. 
    }
    \label{fig:ppf-vs-hit}
\end{figure}

\subsection{Quantitative Evaluation of HiT}
\label{sec:exps:quantitative}

In Fig.~\ref{tab:interpretability-assessment}, we compare the normalized insertion-deletion curves obtained from HiT and other baseline models using the blur strategy, as established in the literature. The baselines assessed include A-ViT~\cite{yin2022avit}, B-cos~\cite{Boehle2024TPAMI}, and DeiT-B~\cite{pmlr-v139-touvron21a}. For B-cos, the saliency maps are intrinsically generated, while for DeiT-B, we rely on post-hoc extraction methods. Specifically, we use the Rollout Matrix~\cite{abnar2020quantifying} and GradCAM~\cite{8237336}, referred to as DeiT-R and DeiT-GC, respectively. To create A-ViT saliency map, we utilized the layer where tokens were discarded, as described in their paper. Additionally, to mitigate the bias of the sorting algorithm when selecting the most important tokens (which tends to favor top-left corner tokens first), we added a small amount of random noise to the saliency map.

Additionally, we evaluate ProtoPFormer~\cite{Xue2022ProtoPFormerCO} on the CUB-2011 dataset. ProtoPFormer employs a Rollout Matrix to filter out irrelevant tokens for its final computation, and we used this Rollout Matrix to define the salient regions for the insertion-deletion analysis.

The results of this experiment are presented in Fig.~\ref{fig:ppf-vs-hit}a for ImageNet and Fig.~\ref{fig:ppf-vs-hit}b for CUB-2011. The nAUC scores and the profiles of the curves indicate that HiT outperforms other ID methods in terms of interpretability.

\begin{table}[t]
    \centering
    \tiny
    \begin{tabular}{C{0.07\textwidth}|C{0.035\textwidth}|C{0.025\textwidth}C{0.025\textwidth}C{0.025\textwidth}C{0.025\textwidth}C{0.025\textwidth}C{0.025\textwidth}}\toprule
       \textbf{Model}   & \textbf{Interp.} & \textbf{IMNet} & \textbf{CUB}     & \textbf{Dogs} & \textbf{Cars}  & \textbf{Aircraft} & \textbf{Pets}\\ \midrule
       DeiT3-B & $\chi$   & 83.6     & 84.9    & 94.0 & 92.8 & 85.3 & 95.0 \\
       DeiT-B  & $\chi$   & 81.1     & 84.9    & 93.4 & 93.0 & 84.9 & 95.1 \\
       B-cos-B & \checkmark & 74.4     & -       & -    & -    & - & - \\
       HiT-B   & \checkmark & 75.0     & 79.0    & 86.8 & 86.2 & 79.8 & 88.6 \\ \midrule
       DeiT3-S & $\chi$   & 81.4     & 83.1    & 90.6 & 93.0 & 83.9 & 94.5  \\
       DeiT-S  & $\chi$   & 79.8     & 83.0    & 89.6 & 92.4 & 83.4 & 94.3\\
       A-ViT-S & \checkmark & 78.6     & -       & -    & - & -& -   \\
       B-cos-S & \checkmark & 69.2     & -       & -    & - & -& -   \\
       ProtoPFormer-S & \checkmark & -        & 84.9    & 90.0 & 90.9 & - & -\\
       HiT-S   & \checkmark & 71.4     & 76.1    & 80.3 & 85.2  & 77.4 & 88.5\\ \bottomrule
    \end{tabular}
    \caption{\textbf{Top1 Accuracy.} Our proposed models have a clear performance loss compared to other ViTs. However, the ViT gains come at the expense of interpretability, whereas the HiT has acceptable performance while being explicable. ProtoPFormer performance were extracted from their paper.}  
    \label{tab:main-results-hit}
    \vspace{-3mm}
\end{table}

To further enhance our understanding of HiT's interpretability, we analyze the unnormalized insertion-deletion metrics of traditional post-hoc methods compared to HiT in Tab.~\ref{tab:interpretability-assessment}. Specifically, we compare the saliency maps extracted from HiT with those generated by GradCAM~\cite{8237336} and an adapted Rollout Matrix~\cite{abnar2020quantifying}, both computed on HiT. Overall, the inherent saliency maps of HiT demonstrate superior performance compared to the post-hoc methods for both insertion and deletion metrics. This finding suggests that these post-hoc algorithms do not consistently identify the regions used for classification.

Finally, in Tab.~\ref{tab:main-results-hit}, we present the accuracy performance of our proposed architecture compared to both non-interpretable and interpretable alternatives. As expected, all interpretable architectures, with the exception of ProtoPFormer, show a decrease in performance relative to the non-interpretable baseline. However, our Hindered Transformer maintains clear advantages in interpretability without a significant drop in performance.

\subsection{Qualitative Evaluation}

\begin{figure}[t]
    \centering
    \includegraphics[width=\linewidth]{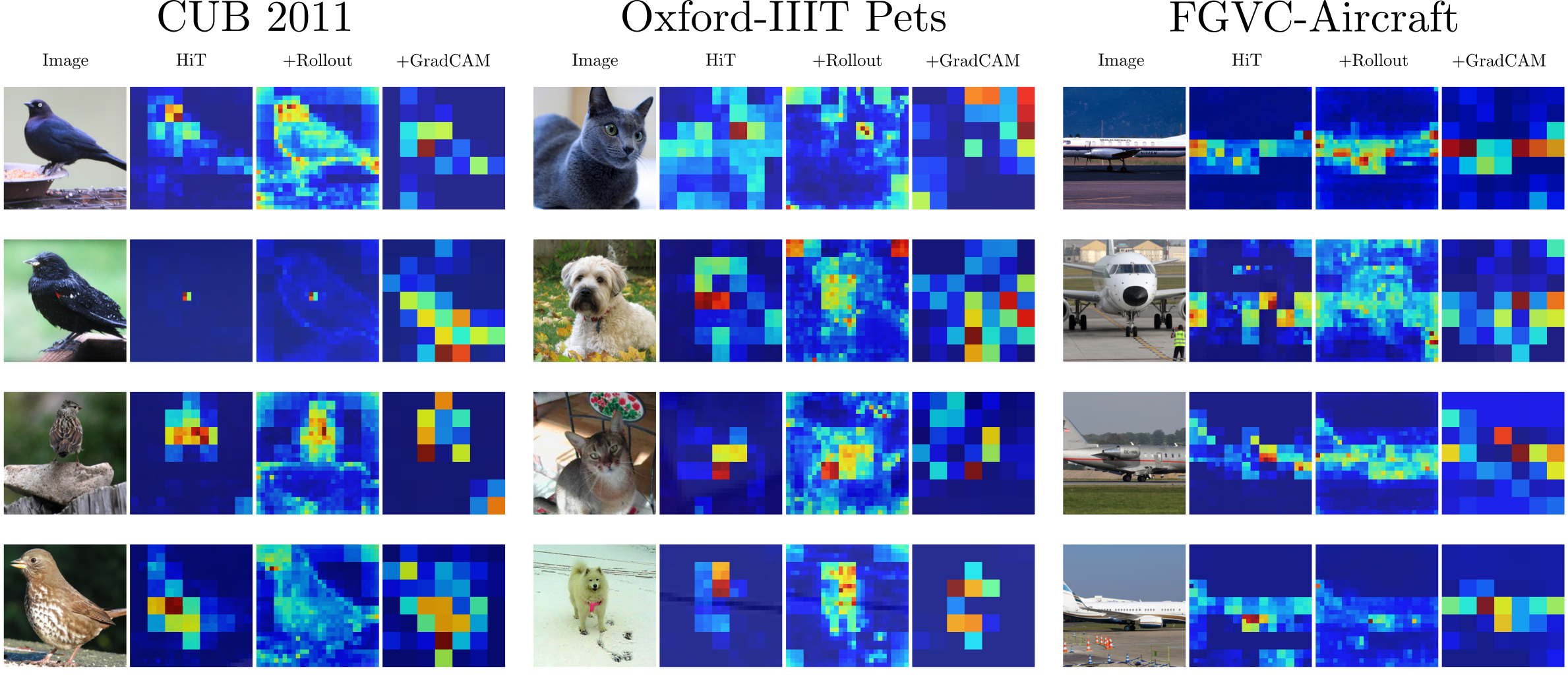}
    \caption{\textbf{Qualitative Comparison.} We show the image and its saliency maps produced by HiT and their homologuous using Rollout and GradCAM. We noticed that HiT tends to use the object's features in the image for its prediction, independently if its prediction is erroneous or not.}
    \label{fig:qualitative-hit}
    \vspace{-3mm}
\end{figure}

In the next part of our study, we show in Fig.~\ref{fig:qualitative-hit} a qualitative comparison between the saliency maps generated by HiT and those computed with GradCAM and the Rollout Matrix when applied to HiT. We make three main conclusions. First, we found that our method focuses on certain parts of objects, regardless of whether the prediction is accurate or not. To quantify our claim, we computed the center of mass for each explanation and checked whether it fell within the object's bounding box in the CUB test set. We achieved 93.4\% accuracy, which strongly supports our claim. Second, our qualitative analysis suggests that misclassification generally occurs due to similar features between image classes. However, since our method generates saliency maps, HiT inherently adopts their weaknesses: it shows where the decision was made, but not which features were used. Third, Rollout and GradCAM produce noisy maps. For example, the former shows edges and highlights the general shape of the object, while the latter produces misaligned coarse maps with our base model.

\subsection{Layer-wise Contribution}

\begin{figure}[t]
\begin{subfigure}{0.495\textwidth}
    \centering
    \includegraphics[width=0.95\textwidth]{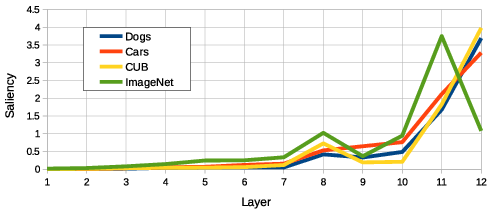}
    \caption{Layer saliencies per dataset.}
\end{subfigure}
\begin{subfigure}{0.495\textwidth}
    \centering
    \includegraphics[width=0.95\textwidth]{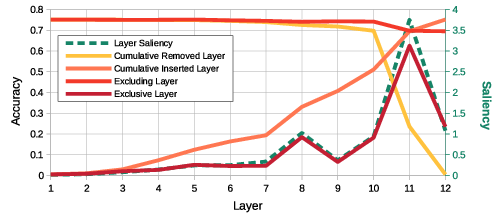}
    \caption{Empirical validation of layer-wise saliency.}
\end{subfigure}
    \caption{\textbf{Layer Saliency.} HiT has more advantages than just image saliency. (a) The first experiment shows that HiT computes the contribution per layer. Without any surprise, the final layers have a greater contribution. (b) We empirically validate our findings in ImageNet with a variety of experiments. Indeed, the results show that by removing certain layers, we obtain larger expected results congruent with the layer saliency.}
    \label{fig:layer-saliency}
    \vspace{-3mm}
\end{figure}

Another advantage of HiT is that we can compute the contribution of each layer. 
Similar to computing the saliency maps spatially, we can create the layer-wise output tokens and look for their individual contributions. 
To this end, we show the results in Fig.~\ref{fig:layer-saliency}a for four tested dataset.
Without any surprise, we can see that most of the discriminative features are in the final layers. 

To ensure that our results are valid, we tested several ablations of our trained model on the ImageNet dataset, shown in Fig.\ref{fig:layer-saliency}b. 
For instance, we tested the accuracy drop by removing or adding a layer of choice (\emph{Excluding/Exclusive Layer} in the figure). 
Similarly, we check the performance loss by removing/adding layers in a cascaded manner, dubbed \emph{cumulative removed/inserted layer}. 
The results corroborate our previous conclusions: our novel architecture is capable of showing the contribution of each individual layer without relying on external methods, such as Linear Probing~\cite{alain2016understanding}, to understand the basic functioning of their inner layers.

\subsection{Sanity Check}

\begin{figure}[t]
\centering
\begin{subfigure}{0.95\linewidth}
    \centering
    \includegraphics[width=0.9\textwidth]{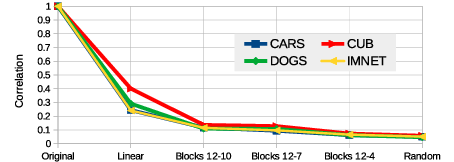}
    \caption{Saliency Rank Correlation}
\end{subfigure}
\hfill
\begin{subfigure}{0.95\linewidth}
    \centering
    \includegraphics[width=0.9\textwidth]{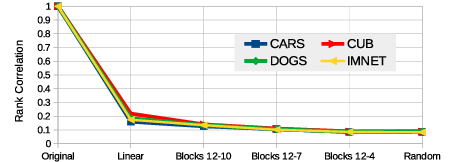}
    \caption{Saliency Correlation}
\end{subfigure}
\begin{subfigure}{0.95\linewidth}
    \centering
    \includegraphics[width=0.9\linewidth]{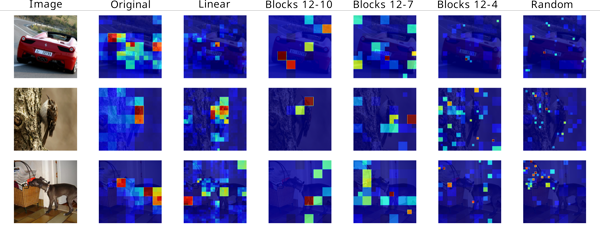}
    \caption{Example Images}
\end{subfigure}
\caption{\textbf{Sanity checks on HiT.} We measure the (a) rank correlation and (b) Pearson correlation of HiT saliency maps before and after randomising the parameters of a layer.   (c) Some visual examples of the saliency maps of randomised models.}
\label{fig:sanity}
\vspace{-3mm}
\end{figure}

Adebayo~\etal~\cite{NEURIPS2018_294a8ed2} highlight that certain maps, such as edge detectors, may appear visually coherent, although they are actually unrelated to the model's decision. 
In order to address this issue, they proposed sanity check, which involves the iterative randomisation of the parameters of the network layers. This process begins with the deepest layers and proceeds step-by-step to the shallower ones. The resulting saliency map produced at each randomization step is then compared with the original map.
In our study, we adopt the same methodology as Adebayo~\etal~\cite{NEURIPS2018_294a8ed2} and compute the absolute rank correlation between the original saliency map and the one generated after randomization, as shown in Fig.~\ref{fig:sanity}a. Additionally, we include the absolute Pearson correlation coefficient in Fig.~\ref{fig:sanity}b, and some examples in Fig.~\ref{fig:sanity}c.

Foremost, we noticed that the rank correlation has a steeper slope than the Pearson correlation for the linear classification layer. 
This shows that there is a greater similarity with the Pearson correlation. 
However, the values are relatively low (less than 0.5), indicating large variations.
Secondly, both metrics reach a plateau for the subsequent randomized models. 
This low similarity suggests that the salient regions highlighted by our model are indeed what the model sees.
Finally, Fig.~\ref{fig:sanity}c shows some qualitative examples produced by the randomization of all blocks, showing that, effectively, the weights' randomization reflect a large variation in the produced saliency.

\subsection{Ablating HiT}\label{sec:perf-loss}

In \S\ref{sec:pooling}, we suggested that token pooling layers will increase the representation power of HiT. 
We hypothesize that the lack of inter-token connections would downgrade greatly the performance. 
Thus, in this section, we empirically validate that the inclusion of token pooling layers increases the performance of the model. 
In addition, we implemented a more powerful pooling strategy used by IdentityFormer~\cite{yu2024metaformer}: a 3×3 convolution with a stride of 2.
We focus on this architecture because it shares similar characteristics with HiT, where each token contains its own information.
Lastly, we theorize that optimizing HiT architectures is challenging. To address this, we adopt an approach similar to DeiT3~\cite{Touvron2022DeiTIR}, training our model for 300 and 600 epochs. Finally, we observed that using binary cross-entropy adversely affects the model's performance, contrary to its effect on DeiT3.

\begin{table}[t]
    \centering
    \tiny
    \begin{tabular}{c|ccc|c} \toprule
       \textbf{Architecture} & \textbf{Pooling}     & \textbf{Loss} & \textbf{Epochs} & \textbf{Val ImageNet}\\ \midrule
       HiT-S        & None        & XE   & 300    & 65.6 \\
       HiT-S        & None        & XE   & 600    & 67.8 \\
       HiT-S        & 2x2 AvgPool & XE   & 300    & 69.3 \\
       HiT-S        & 2x2 AvgPool & XE   & 600    & \textbf{71.4} \\
       HiT-S        & None        & BCE  & 400    & 59.9 \\
       HiT-S        & None        & BCE  & 800    & 62.6 \\ \midrule
       HiT-B        & None        & XE   & 600    & 71.5 \\
       HiT-B        & 2x2 AvgPool & XE   & 600    & \textbf{75.0} \\ \midrule
       HiT-s18      & 2x2 AvgPool & XE   & 300    & 65.6 \\
       HiT-s18      & 2x2 AvgPool & XE   & 600    & 69.3 \\
       HiT-s18      & 3x3 Conv    & XE   & 300    & \textbf{75.9} \\ 
       \bottomrule
    \end{tabular}
    \caption{\textbf{Performance loss ablation.} HiT's performance loss stems from the limited information shared between tokens. Concurrently, the results suggest that HiT's optimization problem is more challenging, as extended training periods lead to more significant performance improvements.}
    \label{tab:perf_loss}
    \vspace{-3mm}
\end{table}

We present the results in Table~\ref{tab:perf_loss}. The findings align with our suspicions: the lack of transferred information between patches significantly reduces the model's accuracy. For instance, by merely adding the convolutional pooling layer of IdentityFormer, we increase the accuracy of a HiT-s18~\cite{yu2024metaformer} from $65\%$ to $75\%$. However, these convolutional layers compromise our model's interpretability by entangling information between tokens. 
Despite this gain in performance, when HiT does not use pooling layers, it's interpretability increases significantly, generating more precise explanations -- please refer to the supplemental material for an empirical comparison. 
In addition, the binary cross entropy reduces its performance. 
Finally, unlike DeiT3 training schemes, our network benefits significantly from increasing the number of epochs. We believe that this result indicates HiT has not yet converged.
\section{Conclusions}



This paper proposed a novel interpretable transformer-like architecture: Hindered Transformer. HiT enhances interpretability by decoupling contributions from individual image patches, enabling the extraction of saliency maps without external tools. 
Extensive experiments across multiple datasets demonstrated the improved interpretability benefits from HiT with a reasonable drop in performance. HiT presents a promising approach, offering favorable trade-offs for applications where interpretability is critical. 

Even though our proposed architecture has many advantages in the terms of interpretability, HiT has limiting factors: slow convergence and the potential challenges in capturing complex dependencies. Regarding the former, we believe that a more comprehensive hyperparameter search, which we were unable to conduct due to limited computational resources, could significantly reduce the training time. Regarding the latter, HiT has some token interactions during the self-attention mechanism, yet, it is indeed weaker than standard ViTs. This is troublesome for spatial tasks and would require substantial modifications, opening opportunities for future research.

\textbf{Acknowledgements}
This work was supported by the Agence Nationale pour la Recherche (ANR) under award number ANR-19-CHIA-0017.

{
    \small
    \bibliographystyle{ieeenat_fullname}
    \bibliography{main}
}

\appendix
\onecolumn
\begin{center}
    {\LARGE \bf Supplementary Material}
\end{center}

\section{Model architecture details}

To compare our architecture as fairly as possible with standard methods (\ie, DeiT~\cite{pmlr-v139-touvron21a}), we follow the same hyperparameter selection for both the base and small versions as their ViT equivalents. That is, HiT base (HiT-B) follows the same architecture as ViT-B. Similarly, HiT small (HiT-S) follows the same hyperparameters as ViT-S. The only differences are the pooling layers, the initial patch size, and that we have removed the last MLP block as it is not used. Tab.~\ref{tab:hyperparameter-conf} shows the architecture hyperparameter choices compared to Visual Transformers.

\begin{table}[h]
    \centering
    \small
    \begin{tabular}{c|cccccc} \toprule
        Version & Layers & Feature dimension & Heads & Pooling layer location & Patch dimension & Parameter count \\ 
        \midrule
        HiT-B  & 12 & 768 & 12 & $[4, 8]$ & 8 & 81.8 \\
        ViT-B  & 12 & 768 & 12 & -        & 16 & 86.9 \\ \midrule
        HiT-S & 12 & 384 &  6 & $[4, 8]$ & 8 & 20.8 \\
        ViT-S & 12 & 384 &  6 & -        & 16 & 22.1 \\ \bottomrule
    \end{tabular}
    \caption{HiT base and small hyperparameter configurations}
    \label{tab:hyperparameter-conf}
\end{table}

\section{Datasets}

As for the dataset, we evaluated HiT on six diverse image classification datasets: 
i) ImageNet~\cite{deng2009imagenet}:  A large-scale dataset with 1.2 million images and 1,000 classes, often used as a benchmark. 
ii) CUB-2011~\cite{WahCUB_200_2011}: A challenging dataset containing 200 bird classes with only 30 training samples per class on average.
iii) Stanford Dogs~\cite{KhoslaYaoJayadevaprakashFeiFei_FGVC2011}:  A dataset with 120 dog classes and 10,000 training and test images.
iv) Stanford Cars~\cite{krause20133d}:  A dataset featuring 196 car classes with 8,100 training and validation examples.
v) FGVC-Aircraft~\cite{maji13fine-grained}: A dataset of 100 airplance classes with 10,000 images.
vi) Oxford-IIIT Pets~\cite{parkhi12a}: a 37 category dataset with roughly 200 images per class.

\section{Insertion-Deletion Curves}

\begin{figure}[h]
    \centering
    \includegraphics[width=\textwidth]{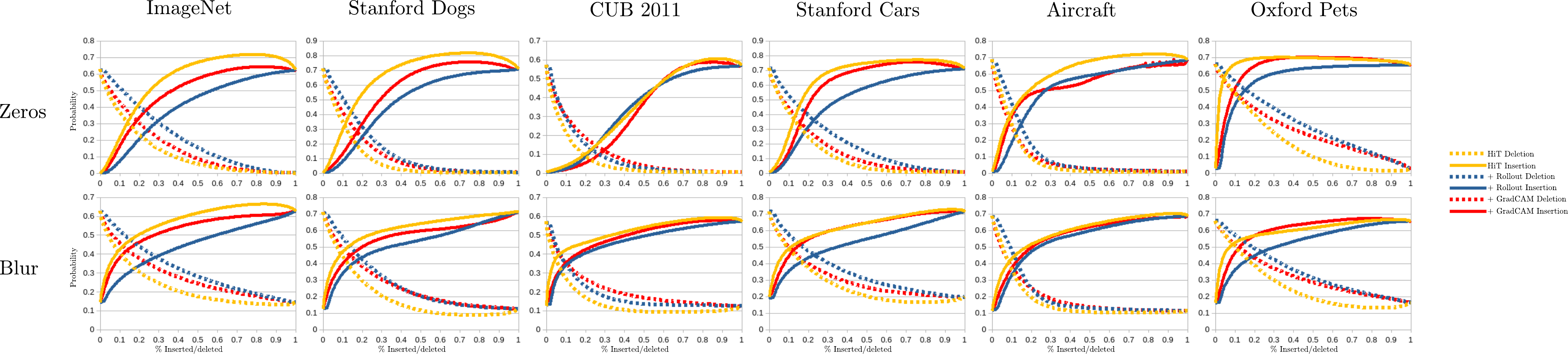}
    \caption{\textbf{Comparing HiT and alternative post-hoc methods.} This experiments reflects the interpretable advantages with respect to traditional post-hoc methods.}
    \label{fig:soup:posthoc-curves}
\end{figure}

In Fig.~\ref{fig:soup:posthoc-curves}, we present the curves from the post-hoc comparison experiment detailed in \S~4.3 of the main document. HiT consistently outperforms both GradCAM~\cite{8237336} and Rollout Matrix~\cite{abnar2020quantifying} across all datasets. Interestigly, GradCAM achieves performance comparable to our method on all datasets except ImageNet~\cite{deng2009imagenet}. We hypothesize that this correlation stems from GradCAM being computed on the final layer tokens, which our analysis shows are the most important (Fig.~5 in the manuscript), except for ImageNet. 

\section{Evaluating HiT without Pooling Layers}

We conducted experiments similar to those in the main manuscript to analyze the positive and negative impact of removing pooling layers. As demonstrated in \S~4.7 of the paper, pooling layers are essential for improving top-1 accuracy performance. However, their inclusion increases the size of the explanations. First, we explore this phenomenon quantitatively in sections \ref{sec:soup:tradeoff} and \S~\ref{sec:soup:layer-contr}. Later, we will explore the qualitative differences in \S~\ref{sec:soup:qual}.

\subsection{Interpretability Trade-off} \label{sec:soup:tradeoff}

\begin{figure}[h]
    \centering
    \includegraphics[width=0.95\linewidth]{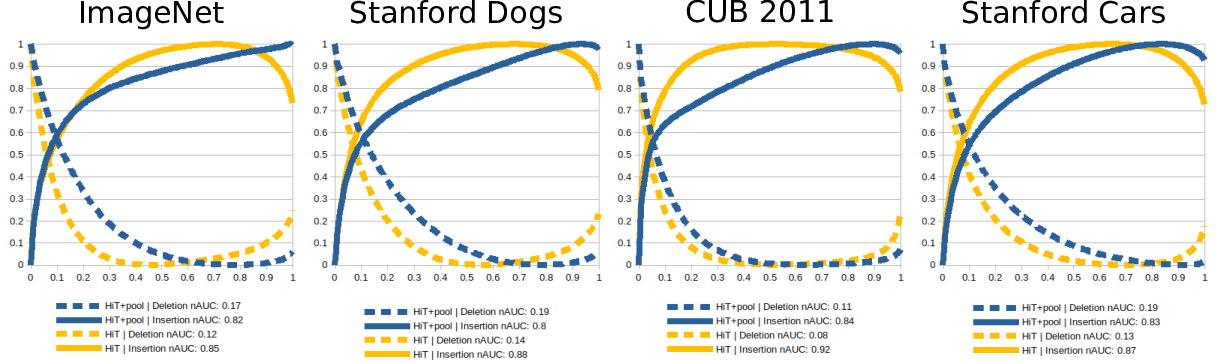}
    \caption{Caption}
    \label{fig:soup:id-pool}
\end{figure}

First, we explore the interpretability gains of HiT without pooling layers. We compare both HiT versions using the normalized insertion-deletion curves on all tested datasets, illustrated in Fig.~\ref{fig:soup:id-pool}. From a quantitative point of view, HiT without any pooling layer is even more interpretable than our proposed architecture. Interestingly, both curves behave similarly, showing a decrease in the insertion probability curve and an increase in the deletion probability curve during their final steps. This is due to the insertion (or deletion) of tokens that adversely affect the model's prediction. 

We also tested GradCAM~\cite{8237336} and the Rollout~\cite{abnar2020quantifying} Matrix directly on our pooling-free HiT architecture, with results shown in Table~\ref{tab:soup:interpretability}. The interpretability gap between post-hoc methods and HiT saliency maps is more pronounced compared to our standard HiT architecture.

\begin{table}[h]
    \centering
    \small
    \begin{tabular}{C{2.2cm}|C{1.25cm}C{1.25cm}|C{1.25cm}C{1.25cm}|C{1.25cm}C{1.25cm}|C{1.25cm}C{1.25cm}}\toprule
        \multirow{2}{*}{Method}       & \multicolumn{2}{c|}{ImageNet} & \multicolumn{2}{c|}{CUB 2011} & \multicolumn{2}{c|}{Stanford Cars} & \multicolumn{2}{c}{Stanford Dogs} \\ \cmidrule{2-9}
                     & Ins-Z (\ua) & Del-Z (\da) & Ins-Z (\ua) & Del-Z (\da) & Ins-Z (\ua) & Del-Z (\da) & Ins-Z (\ua) & Del-Z (\da) \\ \midrule
        HiT           & \textbf{0.65}  &\textbf{ 0.08}  & \textbf{0.56}  & \textbf{0.04}  & \textbf{0.72}  & \textbf{0.05}  & \textbf{0.64}  & \textbf{0.07} \\
        HiT + Rollout & 0.39  & 0.21  & 0.43  & 0.09  & 0.49  & 0.12  & 0.47  & 0.19 \\
        HiT + GradCAM & 0.36  & 0.15  & 0.40  & 0.09  & 0.34  & 0.11  & 0.40  & 0.15 \\ \midrule
                      & Ins-B (\ua) & Del-B (\da) & Ins-B (\ua) & Del-B (\da) & Ins-B (\ua) & Del-B (\da) & Ins-B (\ua) & Del-B (\da) \\ \midrule
        HiT           &  \textbf{0.67} & \textbf{0.16}  & \textbf{0.59}  & \textbf{0.11}  & \textbf{0.65}  & \textbf{0.15}  & \textbf{0.62}  & \textbf{0.18} \\
        HiT + Rollout & 0.47  & 0.31  & 0.50  & 0.22  & 0.50  & 0.29  & 0.50  & 0.32 \\
        HiT + GradCAM & 0.48  & 0.29  & 0.52  & 0.21  & 0.51  & 0.29  & 0.52  & 0.31 \\ \bottomrule
    \end{tabular}
    \caption{\textbf{HiT and Explainability methods:} We quantitatively compare HiT maps and those created by GradCAM and the modified rollout matrix (mean attention). The assessment shows that HiT maps are in fact more faithful to those generated by GradCAM and the rollout matrix. Higher insertion is better, while lower deletion is better. Ins and Del refers to the Insertion and Deletion metrics, respectively. Z is the zero-corrupted image, while B is the blurred corruption strategy.}
    \label{tab:soup:interpretability}
\end{table}

Finally, in Tab.~\ref{tab:soup:accuracy}, we show the performance on the tested datasets. Without any surprise, the loss in performance is major, making it a less appealing option in contrast to our original architecture when computation power is needed. 

\begin{table}[h]
    \centering
    \begin{tabular}{C{0.1\textwidth}|C{0.12\textwidth}|C{0.12\textwidth}C{0.12\textwidth}C{0.12\textwidth}C{0.12\textwidth}}\toprule
        \textbf{Model}   & \textbf{Pooling?} & \textbf{ImageNet} & \textbf{CUB}     & \textbf{Dogs} & \textbf{Cars} \\ \midrule
        \multirow{2}{*}{HiT-S} & $\chi$       & 67.3     & 76.1    & 77.1 & 83.9\\
                               & $\checkmark$ & 71.4     & 76.1    & 80.3 & 85.2 \\ \midrule
        \multirow{2}{*}{HiT-B} & $\chi$       & 71.5     & 76.3    & 80.2 & 84.7\\
                               & $\checkmark$ & 75.0     & 79.0    & 86.8 & 86.2 \\ \bottomrule
    \end{tabular}
    \caption{\textbf{Top1 Accuracy.} Including pooling layers provides a clear advantage in terms of raw performance. However, this performance gain comes at the cost of reduced interpretability.}  
    \label{tab:soup:accuracy}
\end{table}

\subsection{Layer-wise Contributions} \label{sec:soup:layer-contr}

\begin{figure}[h]
    \centering
        \begin{subfigure}{0.49\textwidth}
        \centering
        \includegraphics[width=\textwidth]{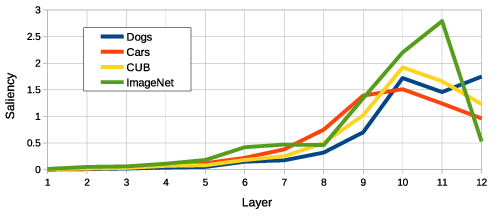}
        \caption{Layer Saliency}
        \label{fig:soup:layer-saliency}
    \end{subfigure}
    \begin{subfigure}{0.49\textwidth}
        \centering
        \includegraphics[width=.95\textwidth]{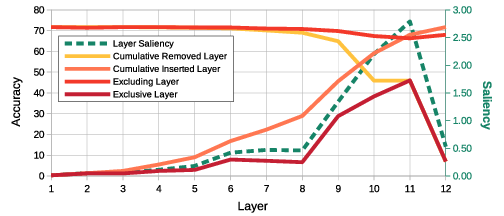}
        \caption{Empirical validation of layer-wise saliency.}
        \label{fig:soup:layer-saliency-exps}
    \end{subfigure}
    \caption{As in the main manuscript, we assess our pooling-free HiT layer contribution. Effectively, HiT can discover the contributions for each layer.}
    \label{fig:my_label}
\end{figure}

Next, we investigate the ability of the pooling-free HiT to analyze layer-wise contributions. Fig.~\ref{fig:soup:layer-saliency} illustrates the layer contribution per dataset, while Fig.~\ref{fig:soup:layer-saliency-exps} plots the ablation results on ImageNet~\cite{deng2009imagenet}. Similar to HiT with layer pooling, the most significant contributions come from the final layers. However, unlike the HiT version with pooling layers, all trained models appear to weight their final predictions equally across the last three layers.

\subsection{Qualitative Comparison} \label{sec:soup:qual}

\begin{figure*}[h]
    \centering
    \includegraphics[width=\linewidth]{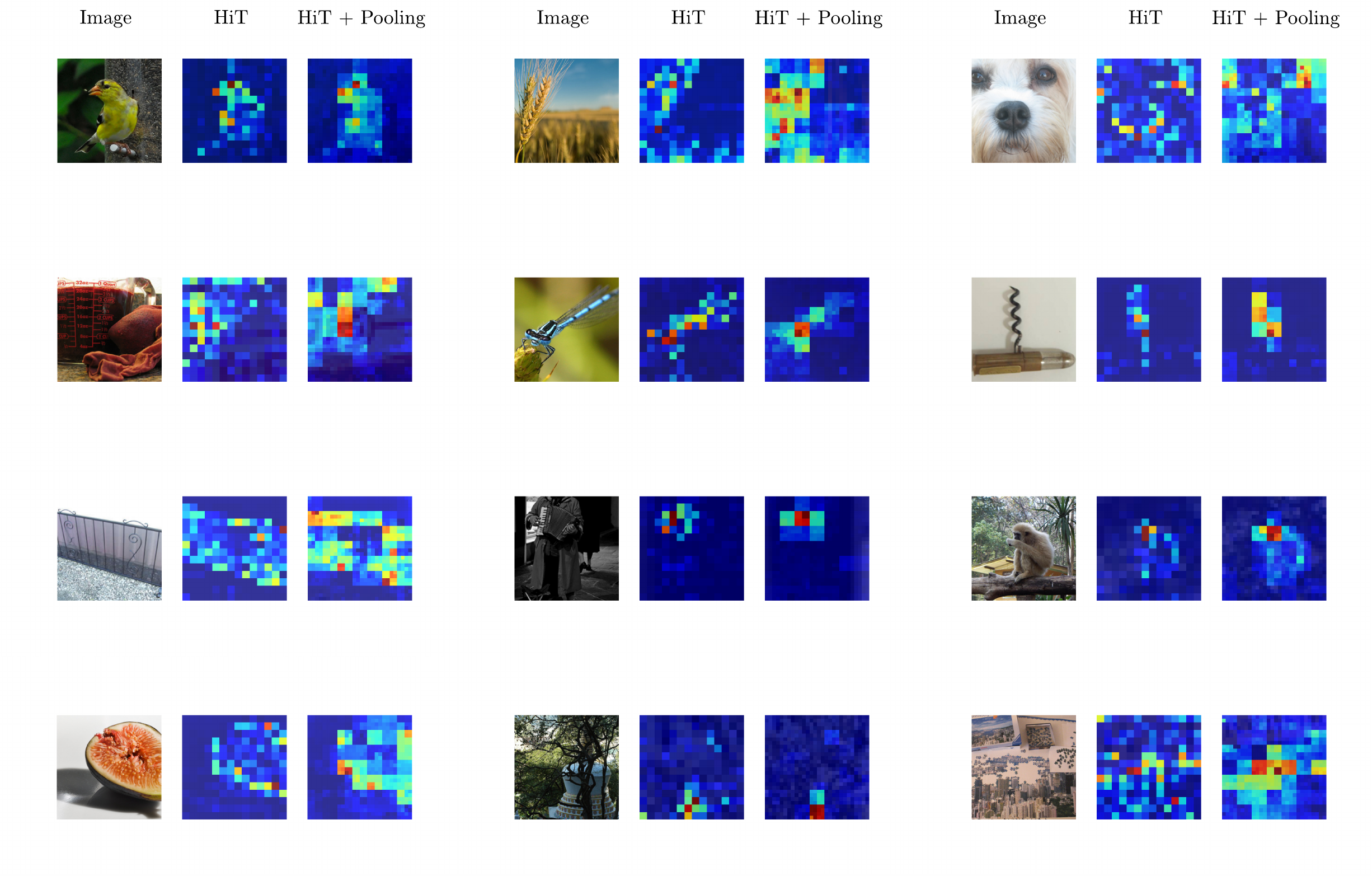}
    \caption{\textbf{Qualitative Examples:} we visually show the difference between HiT saliency maps with and without pooling layers on some correctly classified images from the ImageNet dataset.}
    \label{fig:soup:qualitative}
\end{figure*}

Finally, we qualitatively show the difference between the pool-free HiT and our original version saliency maps in Fig.~\ref{fig:soup:qualitative} in ImageNet~\cite{deng2009imagenet}. As expected, removing the pooling layers produces finer saliency maps.

\end{document}